\documentclass[runningheads]{llncs}
\usepackage[T1]{fontenc}
\usepackage{graphicx,verbatim}
\usepackage{amsmath}
\usepackage{gensymb}
\usepackage{marvosym}
\usepackage{multirow}
\begin{document}
\title{Towards Markerless Intraoperative Tracking of Deformable Spine Tissue}

\author{Connor Daly\inst{1,2}\textsuperscript{(\Letter)} \and
Elettra Marconi\inst{5} \and
Marco Riva\inst{5,6} \and
Jinendra Ekanayake\inst{7} \and
Daniel S. Elson\inst{1,4}
\and
Ferdinando Rodriguez y Baena\inst{1,3}}
%

%
\institute{The Hamlyn Centre for Robotic Surgery, Imperial College London, London, UK, \email{cd1723@ic.ac.uk}\\ 
\and
Department of Computing, Imperial College London, London, UK
\and
Department of Mechanical Engineering, Imperial College London, London, UK
\and
Department of Surgery \& Cancer, Imperial College London, London, UK
 \and
 Department of Neurosurgery, IRCCS Humanitas Research Hospital, Rozzano, Lombardy, Italy
\and
Department of Biomedical Sciences, Humanitas University, Pieve Emanuele, Milan, Italy.
 \and
Stanford University School of Medicine, Stanford University, California, USA
}

\authorrunning{C. Daly \textit{et al.}}

\maketitle              
\begin{abstract}
Consumer-grade RGB-D imaging for intraoperative orthopedic tissue tracking is a promising method with high translational potential. Unlike bone-mounted tracking devices, markerless tracking can reduce operating time and complexity. However, its use has been limited to cadaveric studies. This paper introduces the first real-world clinical RGB-D dataset for spine surgery and develops \textit{SpineAlign}, a system for capturing deformation between preoperative and intraoperative spine states. We also present an intraoperative segmentation network trained on this data and introduce \textit{CorrespondNet}, a multi-task framework for predicting key regions for registration in both intraoperative and preoperative scenes.


\end{abstract}
\section{Introduction and Motivation}
Real-time intraoperative markerless tracking of orthopedic tissue using RGB depth sensors is still in its infancy but holds high translational potential \cite{9646929,liebmann2024automatic}. Current Computer Aided Orthopedic Surgery (CAOS) tracking systems, like the StealthStation S8 (Medtronic, USA), use costly bone-anchored markers to align preoperative and intraoperative data \cite{Medtronic_StealthStation_S8,wilson2024image}. These require additional surgical time for calibration and pose risks of complications \cite{gao2024mazor}. Markerless tracking, instead, uses deep learning-based segmentation to identify relevant points in the surgical scene’s point cloud, aiding registration \cite{9646929,liebmann2024automatic}.

We present the first real-world clinical dataset for training a markerless tracking system, comprising 70,095 colored point clouds from 27 spine surgeries, paired with anonymized preoperative 3D imaging. As a first step toward automatic registration, we train a neural network to isolate key intraoperative points and use it as the backbone of \textit{CorrespondNet}, a model that aligns preoperative and intraoperative data in a shared feature space for registration.

Our contributions include:
\begin{enumerate}
\item The first real-world clinical RGB-D dataset for orthopedic markerless tissue registration.
\item A data labeling system for capturing articulated pose changes in the spine.
\item A markerless tracking architecture that simultaneously segments preoperative and intraoperative data to isolate key regions for registration.
\end{enumerate}

There have been several cadaver-based studies on knee and spine surgery \cite{liebmann2024automatic,chen2021method,9646929,liu2020automatic}. Liu and Rodriguez y Baena \cite{liu2020automatic} pioneered markerless tracking in knee surgery, leading to neural network-based segmentation work by Hu \textit{et al.} \cite{9646929} and Liebmann \textit{et al.} \cite{liebmann2024automatic}. Using a 3D preoperative mesh, conventional registration algorithms such as RANSAC with Iterative Closest Points mapped transformations between the preoperative and intraoperative scenes. Hu \textit{et al.} improved segmentation in occluded scenes, with both Liu and Hu using a PointNet backbone \cite{qi2017pointnet}.

For the spine, Chen \textit{et al.} \cite{chen2021method} registered CT-generated meshes of excised sheep spines against point cloud scans, while Liebmann \textit{et al.} \cite{liebmann2024automatic} applied a similar method to human spine cadavers. Their approach segmented relevant 2D RGB pixels to select depth map points for projection rather than segmenting the entire point cloud. Both studies registered each vertebra separately using ICP.

\begin{figure}
\centering
\includegraphics[width=0.7\textwidth]{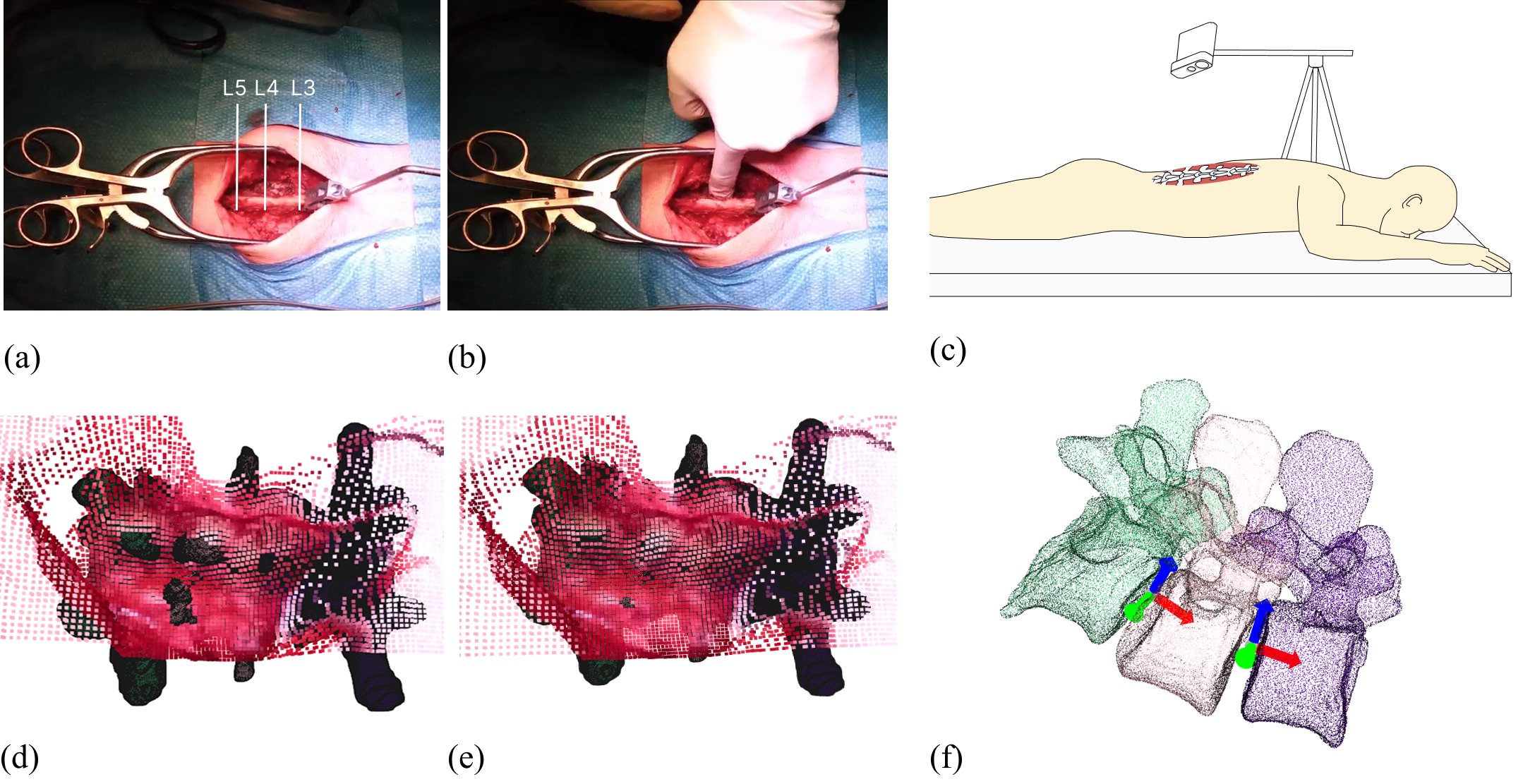}
\caption{\textbf{(a)} RGB images of an exposition site showing L5, L4, and L3 spinous processes. \textbf{(b)} The surgeon indicates the L4 process. \textbf{(c)} Shows the fixed pose RGB-D intraoperative setup. \textbf{(d)} Rigid registration of a preoperative mesh to a surgery scene point cloud. \textbf{(e)} The registered mesh from \textbf{(d)} after deformation in \textit{SpineAlign}. \textbf{(f)} Joint positions and axis vectors for deforming a preoperative CT spine mesh, with blue, red, and green arrows representing anteroposterior, longitudinal, and mediolateral axes.} \label{fig1}
\end{figure}

Despite these advances, real-world surgery presents challenges due to limited tissue exposure. In an open lumbar posterior approach, dissection is typically restricted to the facet joints \cite{vaccaro_posterior_approach}, and key anatomical landmarks remain obscured by residual tissue (Fig.\ref{fig1}(a)). Given the lack of shape cues for vertebral registration, here we first deform the preoperative mesh using a kinematic model before performing global rigid alignment to label our dataset.

For segmentation, following \cite{9646929}, we use a PointNet++ \cite{qi2017pointnet++} backbone. Unlike previous studies however, our network simultaneously segments both the intraoperative scene and preoperative mesh, isolating key regions for registration despite the noisy surgical environment.

This study has full ethics approval. Upon acceptance, the anonymized dataset, labeling code, and model code will be released.
%
%
\section{Methods}
\subsection{Data Collection}
The dataset was collected over a series of 27 spine surgeries, all of which took an open posterior approach. The age of the patients at the time of surgery ranged between 42 and 78 years, with 11 male patients and 16 female patients. The pathologies presented varied; however, the vast majority suffered from spondylolisthesis. 23 patients had accompanying preoperative CT images while the rest had magnetic resonance images (MRI). 24 of the surgeries were lumbar surgeries, 1 was entirely thoracic, whilst the remaining 2 were cervical. In our analysis and inclusion for training our deep learning models, we used the 24 lumbar cases. Within the exposition sites of these surgeries, the number of vertebrae exposed ranges between 2 and 4.
To generate the 3D preoperative state from the acquired CT and MRI data, we incorporated two existing vertebrae segmentation networks. To segment CT scans, we used the `anatomy-consistency' model developed by Meng \textit{et al.}  \cite{meng2023vertebrae}, which is particularly robust to variations in patient spine anatomy. To segment MRI scans, we incorporated the `TotalSpineSeg' network developed by Warszawer \textit{et al.} \cite{warszawer2024fully}, heavily based off `nnU-Net' \cite{isensee2021nnu}. With the segmented scan, we then used the Marching Cubes algorithm \cite{lorensen1998marching}, in conjunction with a simple smoothing filter, to transform the CT representation to a 3D mesh. 

For the intraoperative video sequences (Fig.\ref{fig1}(b)), the Time-of-Flight (TOF) RGB-D camera (Azure Kinect, Microsoft Corporation, USA) was fixed to a static tripod and held over the exposition site via a horizontal boom arm (Fig. \ref{fig1}(c)). Upon completion of the initial exposure by the surgeon, the arm was lowered to a height between 30 - 40 cm from the incision, whereupon we recorded an RGB-D video of approximately 2 minutes length at a frame rate of 30 frames per second. Within the sequence, the operating surgeon pointed to at least three non co-linear `landmarks' of specific vertebrae within the operating site, such as the spinous process and the facet joints. These in-sequence cues were used to provide the gross initial alignment between the preoperative mesh and the intraoperative point cloud scenes, estimating a coarse registration transform via Singular Value Decomposition. 
\subsection{The Spine as a Kinematic Chain}
Since both the camera and patient remain static during surgery, we assume that the most significant intervertebral movement occurs between the preoperative scan (when the patient is supine) and the intraoperative phase (when the patient is prone). Given our focus on the lumbar region, spine movement due to breathing is minimal \cite{donley2022anatomy,leong1999kinematics}. Thus, in the same vein as \cite{liebmann2024automatic}'s system that performs registration with respect to a particular reference frame, for each sequence $s$, we labeled the articulation required to deform the spine from its preoperative state, $Sp$, to its intraoperative state, $Si$, with respect to the point cloud, $P(s_{r})$ from a particular reference frame, $s_{r}$. This frame is typically the first frame in the sequence; however, if there is any obfuscation, this will be the first subsequent frame with a clear view of the exposition.

Given that intervertebral motion in the lumbar region consists primarily of relative rotation \cite{passias2011segmental}, we chose to represent the articulation movement model as a kinematic chain made up of verterbra `links', joined by a ball joint in the center of each intervertebral disc. These joints are parameterized by rotation in three axes: mediolateral, anteroposterior and longitudanal. 
Each intervertebral joint is estimated to be at the midpoint between the centroids of adjacent vertebrae. To determine joint rotation axes, we analyzed the principal components of each vertebra’s shape. For the $i$th joint, the largest component, extending from the vertebra’s centroid to the spinous process, was assigned as the anteroposterior axis. This joint placement can be seen within Fig.\ref{fig1} (f), on a preoperative mesh generated from a CT scan. 

\subsection{Labeling Articulation State}
 With this motion model, for each sequence we labeled the rotation parameters for all joints, as well as an additional global transform, to finely align the preoperative mesh with the reference cloud. In order to determine these parameters, our labeling system used a semi-automated approach. An initial search of the parameter space was performed using basin hopping \cite{wales1997global} in conjunction with the limited-memory Broyden-Fletcher-Goldfarb-Shanno algorithm \cite{nocedal1980updating}. The optimization goal was equally weighted between two terms. The first was a distance-wise correspondence count between a point cloud sampled from the preoperative mesh and the intraoperative point cloud. The second was a `containment' term to encourage the mesh to lie below the surface of the intraoperative point cloud. Among the set of correspondences, for each preoperative point $i$, the vector $v_{i}$ connecting it to its corresponding intraoperative point is compared with the estimated normal $n_{i}$ at that intraoperative point. This is assigned a value and summed as described in Eqs.\ref{eq2} and \ref{eq3}.
We included box constraints within the minimization method, restricting relative vertebra rotation to values consistent with those seen in empirical studies \cite{penning2005measurement,wong2004flexion,wilke2017vitro}. 

\begin{equation}
    \delta_{i} =
\begin{cases} 
0 & \text{if } n_{i} \cdot v_{i} \geq 0, \\
1 & \text{otherwise}
\end{cases}
\label{eq2}
\end{equation}
\begin{equation}
    I = \frac{1}{N}\sum_{i=1}^{N}{\delta_{i}}
\label{eq3}
\end{equation}

We first tested this optimization function on a spine phantom that, after an initial laser scan capture (providing the `preoperative' state), had undergone deformations localized to vertebrae L2-T12 (3 target vertebrae) and L4-L1 (4 target vertebrae). Using a correspondence distance of 8 mm and comparing against a simple rigid ICP registration of the preoperative model, the results in the upper section of Tab. \ref{tab:clinical} show that deformation led to improvements in both the `fitness' score (the ratio of valid correspondences to total points in the intraoperative source) and the RMSE of the correspondences, as well as improved visual alignment. The trial on the phantom represented an `ideal' case wherein vertebrae shape detail was fully exposed within the intraoperative scan, providing preliminary validation of the optimization method along with the intrinsic motion model defined by the kinematic chain. 
To label the clinical dataset, the articulated pose generated by basin-hopping was first verified visually by examining the alignment, assessing its correctness with respect to the surgeon-identified in-sequence landmark, as illustrated in Fig. \ref{fig1} (b). Fine manual adjustments could be made to the articulated pose through the interactive user interface of our labeling system, wherein the user can select specific joints in the kinematic chain and adjust the articulation, or adjust the global transform, to better align with the reference. An example of improved alignment of clinical data with deformation can be seen by comparing Figs. \ref{fig1}(d) and (e).

For our dataset labels, we stored the initialized spine mesh in its original pose for each sequence. We recorded the joint parameters needed for deformation and the rigid transform aligning the original pose to the intraoperative reference frame. We then sampled 30,000 points from the deformed-transformed mesh, and performed ICP, with a correspondence threshold of 10 mm, against the remainder of intraoperative point clouds in the sequence. The results of aligning deformed meshes versus rigid meshes across all sequences in the lumbar focused clinical data can be seen in Tab. \ref{tab:clinical}. Across all categories of `exposure' (number of exposed vertebrae), there was an improvement in fitness scores, while the overall RMSE was similar to that of the rigid registrations. The standard deviation of all metrics and across all exposure levels was lower for alignment with deformation than without, suggesting the alignments found by ICP are more stable for the deformed meshes. For each frame, we stored all points in the intraoperative scene that were within 50 mm of the registered mesh. We also stored the `corresponding' region within the point cloud sample from the mesh surface. The stored scene points and the corresponding mesh points serve as labels for training our segmentation network and \textit{CorrespondNet}.

\begin{table}[ht]
\centering
\caption{Fitness score (ratio of correspondences to total points in source) and RMSE distances (mm), comparing ICP registration of preoperative spine meshes  without deformation (rigid) versus those with deformation. For clinical data, the values presented are the mean across frames along with standard deviations. Exposure is the number of target vertebrae to register against.}
\begin{tabular}{c|c|c|cc|cc}
\hline
\multirow{2}{*}{Source} & \multirow{2}{*}{Exposure} & \multirow{2}{*}{No. of Frames} & \multicolumn{2}{c|}{Rigid} & \multicolumn{2}{c}{Deformed} \\ 
\cline{4-7}
& & & Fitness  & RMSE & Fitness & RMSE \\ 
\hline
Phantom & 3 & - & 0.63  & 4.63 & 0.66  & 4.46    \\
Phantom & 4 & -  & 0.58  & 4.39 & 0.62  & 4.00   \\
\hline
Clinical & 2 & 12,202 & $0.54(\pm0.16)$  & $7.18(\pm0.50)$ & $0.56(\pm0.14)$  & $7.19(\pm0.47)$  \\
Clinical & 3 & 43,757 & $0.57(\pm0.10)$ & $7.25(\pm0.49)$ & $0.59(\pm0.09)$&$7.20(\pm0.41)$  \\
Clinical & 4 & 6,999  & $0.57(\pm0.18)$  & $6.56(\pm0.67)$  & $0.59(\pm0.15)$  & $6.66(\pm0.53)$ \\
\hline
Clinical & Total  & 62,958 & $0.56(\pm0.13)$  & $7.16(\pm0.56)$ & $0.58(\pm0.11)$  & $7.14(\pm0.47)$   \\
\hline
\end{tabular}
\label{tab:clinical}
\end{table}

\subsection{Deep Learning Methods}
Using labeled intraoperative point cloud frames, we first trained a segmentation network, the `pretraining' network, to isolate target points in the surgical scene. The network design was an adaptation of Pointnet++\cite{qi2017pointnet++}, with the six input channels encoding RGB values and spatial coordinates. The architecture is equivalent to the intraoperative path in Fig. \ref{architect}(a). During training, we use an equally weighted sum of the dice loss \cite{dice1945measures} and the focal loss \cite{lin2017focal} as our loss function. The data set was divided into 20 sequences used for training, and 4 sequences as a holdout test set. Hyper-parameter selection, including learning rate and number of training steps, was first validated with a five-fold cross-validation, before training on all 20 sequences. For training all networks, we used an initial learning rate of 0.01, in conjunction with the AdamW optimizer \cite{loshchilov2017decoupled}. Training data augmentation (augmented by $7\times$) included spatial augmentations (rotations and point dropout) and color adjustments.  We defined a training step as a random sample of 10,000 of the augmented training set. We trained the scene segmentation model for 50 steps. All training was done on one Nvidia GeForce RTX 4070 Ti. We report 3 metrics for all models, including the IoU of the predicted point set with respect to the label. To evaluate registration suitability in the absence of a gold standard measurement, we measured shape similarity between source and target point clouds using bidirectional Chamfer and symmetric Hausdorff distances. For scene segmentation, this was between the predicted scene point cloud and its corresponding mesh label.

We then developed a multitask learning network to isolate both the target region within the surgical scene, and the corresponding region within the surface of a deformed unaligned preoperative mesh. The complete architecture of \textit{CorrespondNet} is shown in Fig. \ref{architect}(a). The weights of the `pretraining' network were transferred to the set abstraction layers in \textit{CorrespondNet}, and all layers in the intraoperative path. The preoperative path is similar to the intraoperative path, albeit with the addition of the fusion network. Both paths share the same set abstraction layers. The use of a joint feature space for identifying corresponding regions is novel for such multimodality registration; however, within medical imaging more broadly, it has seen use, for example, in CT to CT registration \cite{grewal2023automatic}.
As input for the mesh segmentation, 30,000 points were sampled from the preoperative mesh surface. We assumed the task of isolating the relevant segments of the mesh depends on the scene context. Accordingly, we fused the final intraoperative feature propagation output with the preoperative feature propagation output. The intraoperative features are first projected to $N\times16$ via a trainable linear layer, max-pooled across channels, concatenated with the mesh features, and finally projected to $M\times8$ using another trainable linear layer. 

Both the segmentation of the mesh and the intraoperative scene use a combined dice-focal loss function. For mesh segmentation, Chamfer and Hausdorff distances are computed between the predicted mesh point cloud and the intraoperative scene label. The network was trained for 100 training steps, with alternating updates with respect to the scene segmentation loss and the mesh segmentation loss. To evaluate the impact of intraoperative information on mesh segmentation, we conducted an ablation study by training \textit{CorrespondNet} without the fusion mechanism.
\section{Results and Discussion}
\begin{figure}[!ht]
\includegraphics[width=\textwidth]{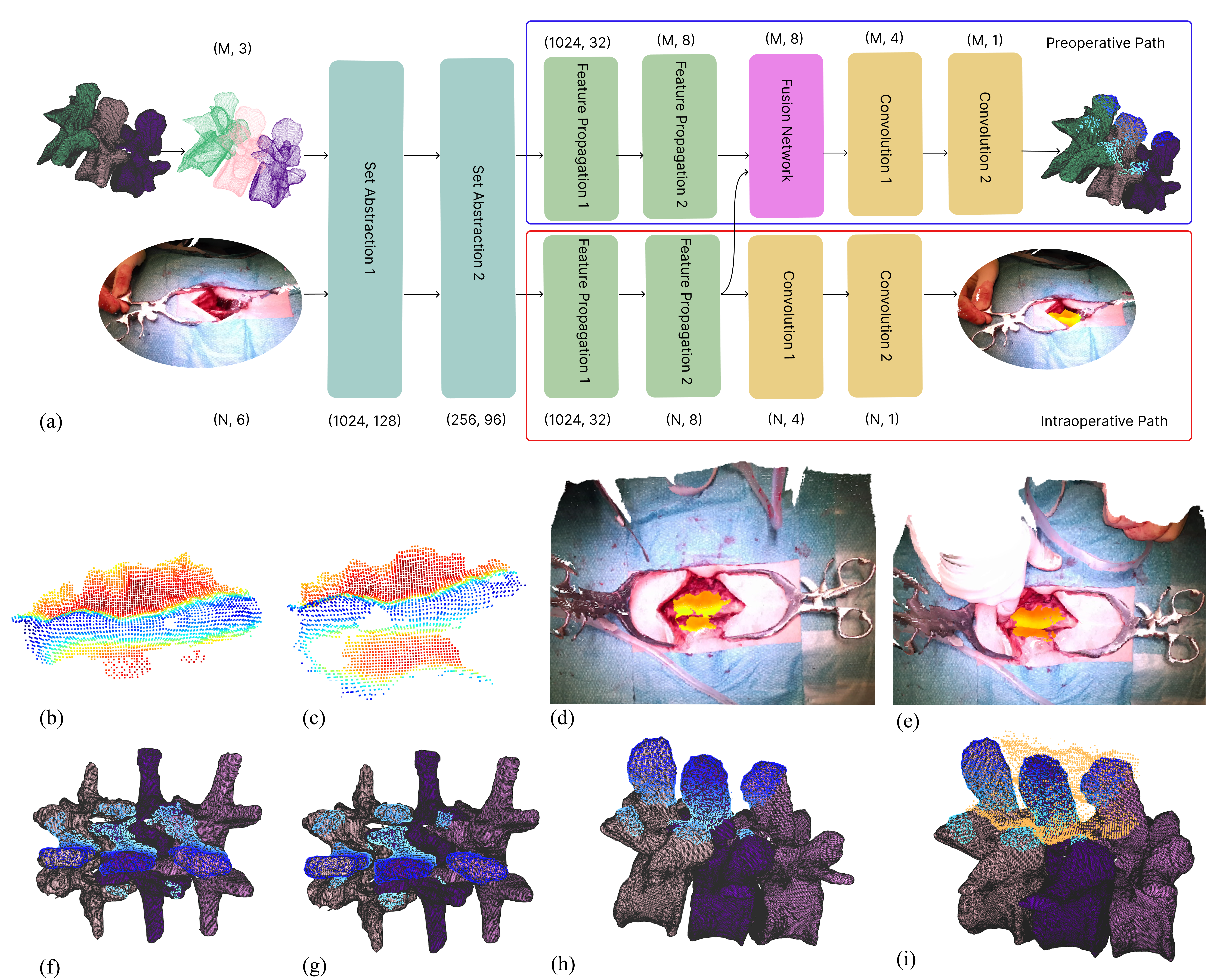}
\caption{\textbf{(a)} Shows the architecture of \textit{CorrespondNet}, where the intraoperative path corresponds to the initial segmentation network. M and N represent the sizes of the preoperative mesh and intraoperative scene point clouds, respectively; \textbf{(b)} and \textbf{(c)} show the scene label and segmentation prediction for a test frame; \textbf{(d)} places the prediction in the scene context; \textbf{(e)} highlights the network’s resilience to surgical scene dynamics, maintaining predictions despite occlusions; \textbf{(f)} displays preoperative mesh and its point cloud label; \textbf{(g)},\textbf{(h)} display preoperative mesh and predicted point cloud; \textbf{(i)} overlays both preoperative and intraoperative predictions on the aligned mesh.}\label{architect}
\end{figure}
\begin{table}[ht]
\centering
\caption{Results for the initial scene segmentation network (PretrainingNet), \textit{CorrespondNet}, and the ‘No Fusion’ ablation study. Intraoperative columns show segmentation results for the intraoperative scene point cloud, while preoperative columns correspond to the preoperative mesh. Values are means, with standard deviation, across all test set frames. Chamfer and Hausdorff distances are in mm.}
\begin{tabular}{l|ccc|ccc}
\hline
\multirow{2}{*}{Network} & \multicolumn{3}{c|}{Intraoperative} & \multicolumn{3}{c}{Preoperative} \\ 
\cline{2-7}
 & IoU  & Chamfer & Hausdorff & IoU & Chamfer & Hausdorff \\ 
\hline
PretrainingNet  & $0.79(\pm0.00)$  & $9.66(\pm0.05)$  & $19.9(\pm0.20)$  & - & -  & - \\
CorrespondNet   & $0.78(\pm0.00)$  & $8.66(\pm0.03)$  & $17.5(\pm0.21)$  & $0.37(\pm0.02)$ & $9.47(\pm0.39)$  & $19.4(\pm1.19)$    \\
No fusion & $0.83(\pm0.00)$  & $8.09(\pm0.02)$  & $19.0(\pm0.19)$  & $0.28(\pm0.04)$ & $13.7(\pm2.15)$  & $28.2(\pm4.27)$    \\
Baseline   & $0.00(\pm0.01)$  & $299(\pm94.9)$  & $180(\pm48.1)$  & $0.01(\pm0.04)$ & $54.7(\pm21.5)$  & $59.0(\pm11.4)$ \\
\hline
\end{tabular}
\label{tab:segmentation}
\end{table}
For each task, we provide baseline metrics in the absence of a gold standard. For intraoperative segmentation, a random neighborhood, equivalent in size to the network prediction, is sampled from the intraoperative scene. For preoperative segmentation, the baseline is sampled from the preoperative mesh. The input for intraoperative segmentation evaluation consists of TOF camera point clouds (500,000–700,000 points) from four test sequences. Preoperative segmentation is evaluated using 30,000 sampled surface points from test set meshes. Results are shown in Tab. \ref{tab:segmentation}. For the single-task segmentation network (row 1, Tab. \ref{tab:segmentation}), all metrics improve over the baseline. Figs. \ref{architect}(b) and (c) illustrate the label-prediction shape difference for a test frame, showing consistent performance despite surgical scene dynamics. Figs. \ref{architect}(d) and (e) confirm robustness against occlusions like the surgeon’s hand. Predictions remain clustered within the exposition site, reflected in low Hausdorff values. \textit{CorrespondNet} (row 2, Tab. \ref{tab:segmentation}) outperforms the baseline for mesh segmentation. Figs. \ref{architect}(f) and (g) show predicted segmentations aligning well with labels, particularly around the spinous processes and facet joints. The intraoperative scene segmentation also improves, reducing Chamfer and Hausdorff distances. Fig. \ref{architect}(i) shows strong overlap in predicted intraoperative and preoperative segmentations. Removing the fusion mechanism (No Fusion) worsens mesh segmentation across all metrics, highlighting the intraoperative scene’s contribution to preoperative segmentation. Both the ablation network and \textit{CorrespondNet} show reduced Chamfer and Hausdorff distances for scene segmentation, likely due to multitask learning effects that improve intraoperative scene segmentation through shared representations \cite{caruana1997multitask}.
\section{Conclusion}
With this work, we took the first step towards developing a real-world markerless pipeline for spine surgery. Our novel architecture, capable of segmenting both preoperative and intraoperative data, could allow for more reliable registration in real-world surgical settings. However, there are limitations in our setup; for example, within data collection, the anatomical landmarks indicated by the surgeon are a relatively coarse cue for label alignment. In a future study, for which we now have ethics approval, a more precise set of alignment indicators will be established within the recording process by integrating intraoperative fluoroscopy.

%
%
%
%
\bibliographystyle{splncs04} 
\bibliography{bib}

\begin{thebibliography}{10}
\providecommand{\url}[1]{\texttt{#1}}
\providecommand{\urlprefix}{URL }
\providecommand{\doi}[1]{https://doi.org/#1}

\bibitem{caruana1997multitask}
Caruana, R.: Multitask learning. Machine Learning  \textbf{28},  41--75 (1997)

\bibitem{chen2021method}
Chen, L., Zhang, X., He, Y., Wang, W., Zhang, F., Sun, L.: A method of 3d-3d multi-stage non-rigid registration of the spine based on binocular structured light. The International Journal of Medical Robotics and Computer Assisted Surgery  \textbf{17}(4),  e2283 (2021)

\bibitem{dice1945measures}
Dice, L.R.: Measures of the amount of ecologic association between species. Ecology  \textbf{26}(3),  297--302 (1945)

\bibitem{donley2022anatomy}
Donley, E.R., Holme, M.R., Loyd, J.W.: Anatomy, thorax, wall movements. In: StatPearls [Internet]. StatPearls Publishing (2022)

\bibitem{gao2024mazor}
Gao, Z., Zhang, X., Xu, Z., Jiang, C., Hu, W., Zhang, H., Hao, D.: Mazor x robot-assisted upper and lower cervical pedicle screw fixation: a case report and literature review. BMC Geriatrics  \textbf{24}(1), ~916 (2024)

\bibitem{grewal2023automatic}
Grewal, M., Wiersma, J., Westerveld, H., Bosman, P.A., Alderliesten, T.: Automatic landmark correspondence detection in medical images with an application to deformable image registration. Journal of Medical Imaging  \textbf{10}(1),  014007--014007 (2023)

\bibitem{9646929}
Hu, X., Nguyen, A., Baena, F.R.y.: Occlusion-robust visual markerless bone tracking for computer-assisted orthopedic surgery. Transactions on Instrumentation and Measurement  \textbf{71},  1--11 (2022). \doi{10.1109/TIM.2021.3134764}

\bibitem{isensee2021nnu}
Isensee, F., Jaeger, P.F., Kohl, S.A., Petersen, J., Maier-Hein, K.H.: nnu-net: a self-configuring method for deep learning-based biomedical image segmentation. Nature Methods  \textbf{18}(2),  203--211 (2021)

\bibitem{leong1999kinematics}
Leong, J., Lu, W., Luk, K., Karlberg, E.: Kinematics of the chest cage and spine during breathing in healthy individuals and in patients with adolescent idiopathic scoliosis. Spine  \textbf{24}(13), ~1310 (1999)

\bibitem{liebmann2024automatic}
Liebmann, F., von Atzigen, M., St{\"u}tz, D., Wolf, J., Zingg, L., Suter, D., Cavalcanti, N.A., Leoty, L., Esfandiari, H., Snedeker, J.G., et~al.: Automatic registration with continuous pose updates for marker-less surgical navigation in spine surgery. Medical Image Analysis  \textbf{91},  103027 (2024)

\bibitem{lin2017focal}
Lin, T.Y., Goyal, P., Girshick, R., He, K., Doll{\'a}r, P.: Focal loss for dense object detection. In: Proceedings of the IEEE International Conference on Computer Vision (2017)

\bibitem{liu2020automatic}
Liu, H., Baena, F.R.Y.: Automatic markerless registration and tracking of the bone for computer-assisted orthopaedic surgery. IEEE Access  \textbf{8},  42010--42020 (2020)

\bibitem{lorensen1998marching}
Lorensen, W.E., Cline, H.E.: Marching cubes: A high resolution 3d surface construction algorithm. Proceedings of the 14th Annual Conference on Computer Graphics and Interactive Techniques pp. 163--169 (1987)

\bibitem{loshchilov2017decoupled}
Loshchilov, I., Hutter, F.: Decoupled weight decay regularization. In: International Conference on Learning Representations (2019)

\bibitem{meng2023vertebrae}
Meng, D., Boyer, E., Pujades, S.: Vertebrae localization, segmentation and identification using a graph optimization and an anatomic consistency cycle. Computerized Medical Imaging and Graphics  \textbf{107},  102235 (2023)

\bibitem{nocedal1980updating}
Nocedal, J.: Updating quasi-newton matrices with limited storage. Mathematics of Computation  \textbf{35}(151),  773--782 (1980)

\bibitem{passias2011segmental}
Passias, P.G., Wang, S., Kozanek, M., Xia, Q., Li, W., Grottkau, B., Wood, K.B., Li, G.: Segmental lumbar rotation in patients with discogenic low back pain during functional weight-bearing activities. The Journal of Bone and Joint Surgery  \textbf{93}(1),  29--37 (2011)

\bibitem{penning2005measurement}
Penning, L., Irwan, R., Oudkerk, M.: Measurement of angular and linear segmental lumbar spine flexion-extension motion by means of image registration. European Spine Journal  \textbf{14},  163--170 (2005)

\bibitem{qi2017pointnet}
Qi, C.R., Su, H., Mo, K., Guibas, L.J.: Pointnet: Deep learning on point sets for 3d classification and segmentation. In: Proceedings of the IEEE Conference on Computer Vision and Pattern Recognition (2017)

\bibitem{qi2017pointnet++}
Qi, C.R., Yi, L., Su, H., Guibas, L.J.: Pointnet++: Deep hierarchical feature learning on point sets in a metric space. Advances in neural information processing systems  \textbf{30} (2017)

\bibitem{Medtronic_StealthStation_S8}
Stealthstation s8 surgical navigation system. \url{https://europe.medtronic.com/xd-en/healthcare-professionals/products/neurological/surgical-navigation-systems/stealthstation/stealthstation-s8.html}, accessed: 2025-02-26

\bibitem{vaccaro_posterior_approach}
Vaccaro, A., Kandziora, F., Fehlings, M., Shanmughanathan, R.: Posterior open approach - midline approach (t1-s1). \url{https://surgeryreference.aofoundation.org}\url{/spine/trauma/thoracolumbar}, accessed: 2025-02-17

\bibitem{wales1997global}
Wales, D.J., Doye, J.P.: Global optimization by basin-hopping and the lowest energy structures of lennard-jones clusters containing up to 110 atoms. The Journal of Physical Chemistry A  \textbf{101}(28),  5111--5116 (1997)

\bibitem{warszawer2024fully}
Warszawer, Y., Molinier, N., Valo{\v{s}}ek, J., Shirbint, E., Benveniste, P.L., Achiron, A., Eshaghi, A., Cohen-Adad, J.: Fully automatic vertebrae and spinal cord segmentation using a hybrid approach combining nnu-net and iterative algorithm. In: Proceedings on CD-ROM-International Society for Magnetic Resonance in Medicine. Scientific Meeting and Exhibition/Proceedings of the International Society for Magnetic Resonance in Medicine, Scientific Meeting and Exhibition (2024)

\bibitem{wilke2017vitro}
Wilke, H.J., Herkommer, A., Werner, K., Liebsch, C.: In vitro analysis of the segmental flexibility of the thoracic spine. PLoS One  \textbf{12}(5),  e0177823 (2017)

\bibitem{wilson2024image}
Wilson~Jr, J.P., Fontenot, L., Stewart, C., Kumbhare, D., Guthikonda, B., Hoang, S.: Image-guided navigation in spine surgery: from historical developments to future perspectives. Journal of Clinical Medicine  \textbf{13}(7), ~2036 (2024)

\bibitem{wong2004flexion}
Wong, K.W., Leong, J.C., Chan, M.K., Luk, K.D., Lu, W.W.: The flexion--extension profile of lumbar spine in 100 healthy volunteers. Spine  \textbf{29}(15),  1636--1641 (2004)

\end{thebibliography}

\end{document}